\documentclass{article}

     \PassOptionsToPackage{numbers, compress}{natbib}

     \usepackage[preprint]{neurips_2025}

\usepackage[utf8]{inputenc} 
\usepackage[T1]{fontenc}    
\usepackage{hyperref}       
\usepackage{url}            
\usepackage{booktabs}       
\usepackage{amsfonts}       
\usepackage{nicefrac}       
\usepackage{microtype}      
\usepackage{xcolor}         

\usepackage{amsmath}
\usepackage{times}
\usepackage{epsfig}
\usepackage{caption}
\usepackage{float}
\usepackage{placeins}
\usepackage{stfloats}
\usepackage{enumitem}
\usepackage{tabularx}
\usepackage{xstring}
\usepackage{multirow}
\usepackage{xspace}
\usepackage[hang,flushmargin]{footmisc}
\usepackage[linesnumbered,ruled]{algorithm2e}
\SetKwInput{KwFunction}{Function}
\usepackage{pifont}
\usepackage{fancyhdr}
\usepackage{wrapfig}
\usepackage[capitalize]{cleveref}

\usepackage{colortbl}

\newlength{\OriginalColumnsep}
\title{Two-Stage Random Alternation Framework for One-Shot Pansharpening}

\author{%
  Haorui~Chen\thanks{Equal contribution.} \\
  University of Electronic Science\\
  and Technology of China\\
  \texttt{hrchen@std.uestc.edu.cn} 
  \And
  Zeyu~Ren\footnotemark[1] \\ 
University of Electronic Science\\
  and Technology of China\\
  \texttt{zeyuren@std.uestc.edu.cn} \\
  \And
  Jiaxuan~Ren \\
  University of Electronic Science\\
  and Technology of China\\
  \texttt{jiaxuan.ren@std.uestc.edu.cn} \\
  \And
  Ran~Ran \\
  University of Electronic Science\\
  and Technology of China\\
  \texttt{ranran@std.uestc.edu.cn} \\
  \AND
  Jinliang~Shao \\
  University of Electronic Science\\
  and Technology of China\\
  \texttt{shaojinliang@std.uestc.edu.cn} \\
  \And
  Jie~Huang \\
  University of Electronic Science\\
  and Technology of China\\
  \texttt{huangjie@std.uestc.edu.cn} \\
  \And
  Liangjian~Deng\thanks{Corresponding author.}\\
  University of Electronic Science\\
  and Technology of China\\
  \texttt{liangjian.deng@uestc.edu.cn}
}

\begin{document}

\maketitle

\begin{abstract}
Deep learning has substantially advanced pansharpening, achieving impressive fusion quality. However, a prevalent limitation is that conventional deep learning models, which typically rely on training datasets, often exhibit suboptimal generalization to unseen real-world image pairs. This restricts their practical utility when faced with real-world scenarios not included in the training datasets. To overcome this, we introduce a two-stage random alternating framework (TRA-PAN) that performs instance-specific optimization for any given Multispectral(MS)/Panchromatic(PAN) pair, ensuring robust and high-quality fusion. TRA-PAN effectively integrates strong supervision constraints from reduced-resolution images with the physical characteristics of the full-resolution images. The first stage introduces a pre-training procedure, which includes Degradation-Aware Modeling (DAM) to capture spectral degradation mappings, alongside a warm-up procedure designed to reduce training time and mitigate the adverse effects of reduced-resolution data. The second stage employs Random Alternation Optimization (RAO), randomly alternating between reduced- and full-resolution images to refine the fusion model progressively. This adaptive, per-instance optimization strategy, operating in a one-shot manner for each MS/PAN pair, yields superior high-resolution multispectral images. Experimental results demonstrate that TRA-PAN outperforms state-of-the-art (SOTA) methods in quantitative metrics and visual quality in real-world scenarios, underscoring its enhanced practical applicability and robustness.
\end{abstract}
\section{Introduction}
\label{sec:intro}
High-resolution multispectral (HRMS) imagery is crucial\label{checklist:impact} for many remote sensing applications~\cite{yang2023cross, zhang2021unsupervised, meng2019review}. However, due to satellite sensor limitations, multispectral and panchromatic images are usually captured separately~\cite{he2018panchromatic}. Multispectral (MS) images are used for spectral information, while panchromatic (PAN) images provide spatial detail. Pansharpening aims to fuse these two types of images to generate an HRMS image, a process vital for maximizing the utility of satellite data.

Pansharpening techniques have evolved from traditional methods to deep learning (DL) approaches~\cite{DDcGAN, xu2020u2fusion, cao2021pancsc, yang2023panflownet, rui2023unsupervised, yang2023multi}. Traditional methods like component substitution (CS)~\cite{choiNewAdaptiveComponentSubstitutionBased2011,vivoneRobustBandDependentSpatialDetail2019}, multi-resolution analysis (MRA)~\cite{vivoneContrastErrorBasedFusion2014,vivoneFullScaleRegressionBased2018}, and variational optimization (VO)~\cite{fuVariationalPanSharpeningLocal2019,tianVariationalPansharpeningExploiting2022,fu2019LocalGradientConstraints} rely on a solid mathematical foundation to merge spatial and spectral information without requiring training datasets. However, these methods often face spectral distortion and spatial misalignment. In contrast, DL-based techniques, supported by advances in hardware and software, offer improved preservation of spectral and spatial features in high-resolution images.

Many DL-based pansharpening models are developed under a ``train-then-infer'' paradigm, where models are trained on large-scale, often simulated, datasets and then applied to new, unseen images. 
\begin{figure}[t]
    \centering
    \includegraphics[width=\textwidth]{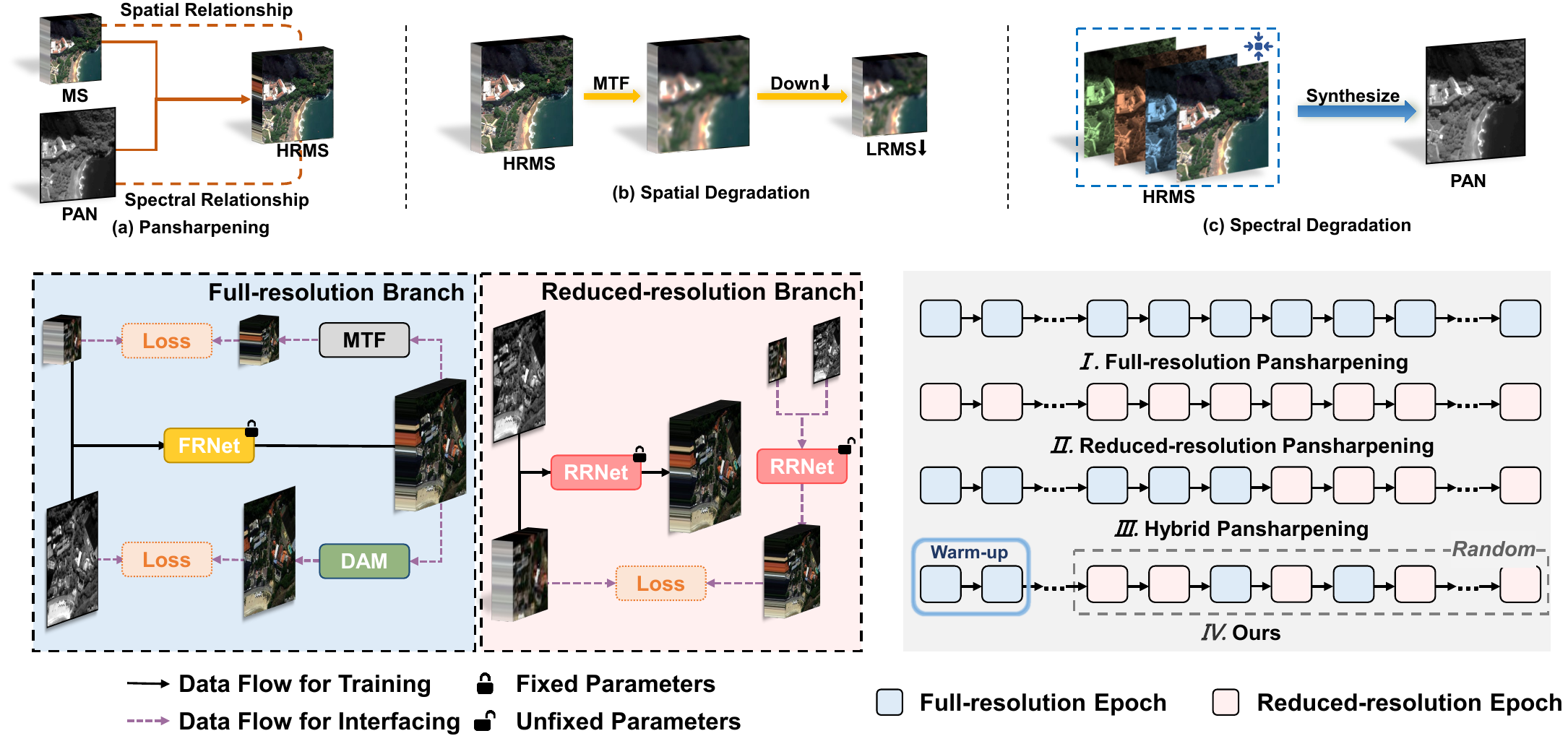}
    \caption{Overview of full-resolution training for pansharpening and different paradigms. The upper panel illustrates the pansharpening task, which includes: \textbf{(a)} a depiction of the fundamental spectral and spatial relationships that link MS, PAN, and HRMS images; \textbf{(b)} the spatial degradation pathway from HRMS to low-resolution multispectral (LRMS) images, a process involving the Modulation Transfer Function (MTF) and downsampling; and \textbf{(c)} a demonstration of how spectral relationships between HRMS and PAN are learned through the synthesis of channels within the HRMS (such as the RGB channels shown), thereby illustrating spectral degradation. The lower panel delineates various pansharpening paradigms: \textbf{(\MakeUppercase{\romannumeral 1})} a full-resolution training approach; \textbf{(\MakeUppercase{\romannumeral 2})} a reduced-resolution training approach; \textbf{(\MakeUppercase{\romannumeral 3})} a hybrid training approach; and \textbf{(\MakeUppercase{\romannumeral 4})} the proposed TRA-PAN method, which incorporates warm-up and random alternating training stages.}
    \label{fig:head}
    \vspace{-16pt}
\end{figure}
However, this presents a critical limitation in practical applications, as these models frequently exhibit a performance decline when applied to real-world MS/PAN image pairs that deviate from the training data distribution. This generalization gap occurs because a single pre-trained model inherently struggles to adapt to the vast diversity of atmospheric conditions and scene content in real-world scenarios. Consequently, their fusion quality on unseen, full-resolution inputs can be unsatisfactory.

To mitigate this generalization challenge and enable models to adapt to specific input characteristics, the concept of one-shot learning (some papers also refer to this as zero-shot learning, the relationship we will elaborate on in \cref{sec:related}) has gained traction in pansharpening~\cite{Zero-shot}. The model is optimized for each MS/PAN pair in this setting. Researchers have typically adapted established training strategies for this one-shot purpose. These adaptations, however, face distinct challenges. One common strategy is unsupervised training directly on the full-resolution images~\cite{rs17010016, rs12142318}. This strategy models the degradation process from an HRMS image. Spatial degradation to the MS image is modeled using MTF filtering and downsampling, as illustrated in \cref{fig:head} (a). The spectral degradation to the PAN image is often modeled via channel synthesis (e.g., linear degradation~\cite{wang2014fusion, li2013remote}), as shown in \cref{fig:head} (b). The overall unsupervised training process, illustrated in \cref{fig:head} (c), enables training without explicit labels. The paradigm for this unsupervised full-resolution strategy is shown in \cref{fig:head} (\MakeUppercase{\romannumeral 1}). \textit{Limitation 1: When applied in a one-shot manner, this strategy's oversimplified imaging process can cause the reconstructed image's spectral response to deviate from the real data.} Another strategy is supervised training using synthetic data derived from the input pair itself, where the original MS image serves as ground truth after degrading both MS and PAN inputs, a paradigm detailed in \cref{fig:head} (\MakeUppercase{\romannumeral 2}). \textit{Limitation 2: This strategy can be constrained by distributional discrepancies between the simulated data and the real-world fusion task, and often relies on a ``scale-invariance" assumption~\cite{ciotola2022pansharpening, gong2024cross, ciotola2024hyperspectral} that may not always hold.} Given these limitations, hybrid training~\cite{Zero-shot, yang2023cross, vitale2020detail, cao2025respandiff}, exemplified by the paradigm in \cref{fig:head} (\MakeUppercase{\romannumeral 3}), attempts to combine strengths. \textit{Limitation 3: However, if hybrid strategies treat resolutions separately during one-shot optimization, they often suffer from ineffective information transfer and fusion.} As a result, the model may still focus on optimizing objectives for each scale somewhat independently, preventing a global optimization for the given instance.

This paper introduces the Two-Stage Random Alternation Framework for Pansharpening (TRA-PAN) to effectively overcome these persistent limitations, representing a paradigm shift towards robust instance-specific optimization. TRA-PAN performs instance-specific optimization for each MS/PAN input pair in a one-shot manner, as illustrated in \cref{fig:head} (\MakeUppercase{\romannumeral 4}). Unlike the approaches mentioned above, it moves beyond learning a universal mapping from a fixed dataset or relying solely on simplified self-supervision derived from either full or reduced resolutions alone. Instead, TRA-PAN dynamically adapts its fusion process based on the unique characteristics of the currently provided MS and PAN images. This ensures an HRMS image tailored to the input, delivering robust, high-quality results across diverse real-world conditions. Furthermore, its unique optimization strategy avoids the pitfalls of resolution separation seen in simpler hybrid methods, leading to better global optimization and achieving generalization for unseen pairs akin to traditional methods. This one-shot, instance-adaptive nature offers significant advantages in practical scenarios where image characteristics vary widely.

TRA-PAN achieves this instance-specific optimization through two independent stages: pre-training and random alternating optimization (RAO). The first pre-training stage encompasses two key procedures: Degradation-Aware Modeling (DAM) and a warm-up procedure. In the DAM procedure, we capture the complex spectral degradation relationship inherent to the input MS/PAN pair, moving beyond generic assumptions. Concurrently, the warm-up procedure only utilizes the full-resolution images to provide the fusion model with excellent initial parameter values. This procedure accelerates convergence and avoids potential issues from early reduced-resolution training, differing from traditional fine-tuning methods that typically pre-train on reduced-resolution images before fine-tuning using full-resolution images~\cite{rs17010016, rs12142318, ciotola2022pansharpening}. Subsequently, the RAO stage leverages the degradation model learned in DAM and the initialized fusion model from the warm-up. RAO employs a dynamic, random alternating training strategy between the reduced- and full-resolution images with a certain probability. This approach effectively combines strong supervision constraints from reduced-resolution with physical characteristics guidance from full-resolution, acting as a regularizer. By balancing these resolutions, TRA-PAN prevents overfitting to a single one and guides the model towards a better optimum, improving the quality of HRMS generation. The main contributions of this work are:

\begin{itemize}
\vspace{-4pt}
\item Our framework optimizes instance-specifically by leveraging both reduced- and full-resolution images, combining their strengths. The random alternating strategy enhances generalization in the one-shot setting by balancing strong and weak supervision across resolutions. Additionally, the warm-up procedure accelerates the training process and mitigates the drawbacks of early reduced-resolution training, breaking away from traditional fine-tuning approaches.

\item We propose TRA-PAN, a novel one-shot, instance-specific optimization framework for pansharpening. It dynamically adapts to individual MS/PAN pairs, ensuring high-quality fusion by learning from the input data itself. This strategy offers significant advantages for practical applications, particularly in scenarios with diverse real-world images deviated from the training data distribution.

\item TRA-PAN demonstrates superior performance over state-of-the-art (SOTA) methods in quantitative metrics and visual quality on real-world full-resolution datasets. This highlights its strong practical applicability and robustness, offering a reliable solution for diverse pansharpening tasks. Furthermore, the framework is designed to be adaptable to various network backbones.
\end{itemize}

\section{Related Work}
\label{sec:related}
\subsection{Resolution-Utilization strategies}
Recent DL-based pansharpening utilizes three main resolution-utilization strategies. Reduced-resolution training, pioneered by \citet{masi2016pansharpening}, downscales MS and PAN inputs, using the original MS for supervision. Several specific methods have been proposed to enhance this approach's performance further. PNN~\cite{PNN} applies a simple three-layer fully-convolutional model. \citet{wei2017boosting} introduces DRPNN, which employs residual learning to construct an intense convolutional neural network, aiming to exploit better the high non-linearity of deep learning models for improved fusion accuracy. \citet{LAGConv} introduces LAGConv, a novel convolution operation that adapts kernels based on local image context. However, these strategies can be constrained by distributional discrepancies between the simulated reduced-resolution images and the real-world full-resolution images.

Full-resolution training directly utilizes MS and PAN images. It is often self-supervised and requires careful loss function design. This approach models the degradation from an HRMS to the observed MS and PAN inputs. This typically involves MTF filters for spatial degradation and techniques like channel synthesis for spectral degradation. An early method in this category is the detail-preserving cross-scale learning strategy, which uses an MTF-GLP-HPM~\cite{wang2021multiresolution} model-based algorithm as a proxy for high-resolution ground truth. Another method, GTP-PNet~\cite{zhang2021gtp}, incorporates gradient transformation to improve spatial and spectral accuracy. Pan-GAN~\cite{ma2020pan}, an unsupervised generative adversarial network framework, effectively preserves both spectral and spatial information by using two discriminators.

Hybrid training combines reduced and full-resolution strategies to leverage their respective strengths, as seen in Cross-Resolution Semi-Supervised Adversarial Learning~\cite{yang2023cross} and CrossDiff~\cite{xing2024crossdiff}. However, separating these into distinct stages might not fully exploit the model's potential.

In contrast, our proposed TRA-PAN framework effectively integrates strong supervision from reduced-resolution images with the physical characteristics of full-resolution images through a novel random alternation optimization, addressing the limitations of these conventional strategies.

\subsection{One-shot learning}
In the broader machine learning domain, one-shot learning generally describes a model's ability to learn a new concept or perform a task from a single or very few examples. Applying this principle to pansharpening, one-shot learning refers to training and testing a model on a single input MS/PAN pair, optimizing the model for that instance. It is worth noting that some prior research in this field has referred to this instance-specific optimization approach as ``zero-shot learning''. Both terms describe the same core concept, but we consider ``one-shot learning'' a more precise descriptor for this per-instance adaptation process.

Pioneering work in applying such internal learning principles to image restoration was conducted by \citet{shocher2018zero} for image super-resolution. Their approach relied on a lightweight network and cross-scale patch matching, demonstrating that task-specific knowledge could be effectively learned from the test image. Building upon these foundations, \citet{Zero-Sharpen} introduced the Zero-Sharpen framework. This method combines deep learning with variational optimization to address scale-variance issues encountered in pansharpening. Furthermore, \citet{Zero-shot} introduced ZS-Pan, a zero-shot semi-supervised method with three stages. \citet{rui2024variational} proposed a variational zero-shot method that employs a neural network to predict the coefficient tensor governing HRMS-PAN relationships. Their approach optimizes this coefficient and the HRMS image within a unified framework, incorporating DIP-type regularization and alternating minimization. These methods exemplify the trend towards dynamic, instance-specific models. Our TRA-PAN advances this paradigm with a two-stage approach, ensuring robust, high-quality fusion tailored to each unique image pair.
\vspace{-4pt}
\section{Method}
\vspace{-2pt}
\label{sec:method}
\subsection{Notation}
\vspace{-2pt}
This subsection introduces key notation. Let \(\mathcal{P} \in \mathbb{R}^{H \times W}\) represent the PAN image, where \(H\) and \(W\) denote height and width, respectively. The MS image, denoted as \(\mathcal{M} \in \mathbb{R}^{h \times w \times c}\), possesses height \(h\), width \(w\), and spectral bands \(c\). The HRMS image \(\mathcal{H} \in \mathbb{R}^{H \times W \times c}\) is obtained by fusing the PAN and MS images. Furthermore, the low-resolution multispectral image (LRMS) is \(\mathcal L \in \mathbb{R}^{\frac{h}{r} \times \frac{w}{r} \times c}\), and the low-resolution panchromatic image (LRPAN) is \(\mathcal{L}^P \in \mathbb{R}^{\frac{H}{r} \times \frac{W}{r}}\), where \(r\) is the resolution ratio between the MS and PAN images, i.e. \(r = \frac{H}{h}\). \textit{These downsampled images are obtained using the MTF filter and the downsampling operation described in the supplementary material.} These degradation processes are represented as follows:
\begin{equation}
\mathcal L, \mathcal{L}^P = \text{MTF}_\downarrow(\mathcal{M}, \mathcal{P})
\end{equation}
where \(\text{MTF}_\downarrow(\cdot)\) symbolizes the combined MTF filtering and downsampling operations.

\vspace{-2pt}
\subsection{Overview}
This section will use the \cref{alg:framework} to outline our TRA-PAN. Our method begins with the pre-training stage, consisting of the DAM and warm-up procedures. They are delineated by \cref{alg:line1}-\cref{alg:line4} and \cref{alg:line7}-\cref{alg:line12}, respectively. 
The DAM procedure trains a spectral degradation-aware network, parameterized by \(\theta_D\), to model the spectral degradation from \(\mathcal{M}\) to \(\mathcal{L}^P\). 
\setlength{\columnsep}{18pt}
\begin{wrapfigure}{r}{0.5\textwidth}
\vspace{-4pt}
\begin{algorithm}[H]
\caption{Overall TRA-PAN Framework}
\label{alg:framework}
\KwData{MS \(\mathcal{M}\), PAN \(\mathcal{P}\) and their \(i\)-th augmentation \(\mathcal{M}_i\), \(\mathcal{P}_i\), DAM training epochs \(n_1\), RAO training epochs \(n_2\), Ratio \(p\), Warm-up epochs \(m\).}
\KwFunction{DAM procedure with parameters \(g(\cdot; \theta_D)\), fusion network in RAO stage with parameters \(f(\cdot, \cdot; \theta_{R})\), combined MTF filtering and downsampling operations \(\text{MTF}_\downarrow(\cdot)\)}
\KwResult{HRMS \(\mathcal{H}\).}

\tcp{Training of DAM procedure}
\For{\(i = 1\) \textbf{to} \(n_1\)}{ \label{alg:line1}
   \( \mathcal{D}^M_i \leftarrow g(\mathcal{M}_i; \theta_D)\)\\
Update  \(\theta_{D} \) based on \cref{eq:DAM}
} \label{alg:line4}
\BlankLine

\tcp{Training of RAO stage}
\For{\(j= 1\) \textbf{to} \(n_2\)}{ \label{alg:line5}
    \(u \sim Uniform(0, 1)\)
    \BlankLine
    
    \tcp{Warm-up procedure}
    \If{\(j > m\)}{ \label{alg:line7}
        \(\mathcal{L}_i \leftarrow \text{MTF}_\downarrow(\mathcal{M}_i)\)\\
        \(\mathcal{L}^P_i \leftarrow \text{MTF}_\downarrow(\mathcal{P}_i)\)\\
        \(\mathcal{M}^H_i \leftarrow f(\mathcal{L}_i, \mathcal{L}^P_i; \theta_{R})\)\\
      Update  \(\theta_{R} \) based on \cref{eq:reduced}
    } \label{alg:line12}
    \BlankLine
    
    \If{\(u < p\)}{ \label{alg:line13}
        \(\widehat{\mathcal{H}} \leftarrow  f(\mathcal{M}, \mathcal{P}; \theta_{R})\)\\
        \(\mathcal{D}^H \leftarrow g(\widehat{\mathcal{H}}; \theta_D)\)\\
        \(\widehat{\mathcal{L}} \leftarrow \text{MTF}_\downarrow(\widehat{\mathcal{H}})\)\\
      Update \(\theta_{R}\)  based on \cref{eq:full}
    } \label{alg:line18}
} \label{alg:line19}
\BlankLine

\tcp{Inference}
\(\mathcal{H} \leftarrow f(\mathcal{M}, \mathcal{P}; \theta_R)\)
\end{algorithm}
\vspace{-10pt}
\end{wrapfigure}
This learned degradation subsequently guides the training of the fusion network in the Random Alternation Optimization (RAO) stage. This strategy accelerates network convergence and mitigates potential negative influences from reduced-resolution training. It provides effective initial parameters \(\theta_R\) for the fusion network, establishing a solid foundation for the subsequent random alternating training. The method then proceeds to the RAO stage, detailed in \cref{alg:line5}-\cref{alg:line19}. Aided by the DAM procedure, \(\theta_R\) is optimized through random alternating training on both reduced- and full-resolution images. Following optimization, the fusion network performs the final inference, taking \(\mathcal{M}\) and \(\mathcal{P}\) as inputs to generate the HRMS image \(\mathcal{H}\).
\begin{equation}
\mathcal{H} = f(\mathcal{M}, \mathcal{P}; \theta_{R})
\end{equation}
Here, \(f(\cdot, \cdot; \theta_{R})\) denotes the fusion network with its learnable parameters.
\subsection{Stage 1: Pre-training}
\paragraph{DAM Procedure} This procedure learns the complex non-linear spectral degradation from \(\mathcal{M}\) to \(\mathcal{L}^P\). Given data scarcity in one-shot training, which can lead to overfitting and protracted training with complex architectures, we utilize a simple yet effective MLP to model spectral degradation. \textit{For a detailed analysis of this MLP architecture, refer to the supplementary material.}

During training, a ten-crop data augmentation is applied to \(\mathcal{M}\), \(\mathcal{L}\), and \(\mathcal{L}^P\). Five cropped sections, with horizontal flips, yield ten augmented patches. This strategy mitigates potential overfitting in the one-shot learning context. \textit{Further details about the data augmentation procedure can be found in the supplementary material.} The data augmentation process can be expressed as:
\begin{equation}
\mathcal{M}_i = \text{DA}(\mathcal{M}), \mathcal{L}_i = \text{DA}(\mathcal{L}), \mathcal{L}_i^P  = \text{DA}(\mathcal{L}^P)
\end{equation}
where \(\text{DA}(\cdot)\) is data augmentation, yielding \(i\)-th augmented MS \(\mathcal{M}_i\), LRMS \(\mathcal{L}_i\), and LRPAN \(\mathcal{L}_i^P\) images. The MLP processes reshaped \(\mathcal{M}_i\) into an intermediate vector \(\widehat{\mathcal{D}_i^M} \in \mathbb{R}^{hw \times 1}\), then reshaped to final degraded MS image \(\mathcal{D}_i^M \in \mathbb{R}^{h \times w}\). This procedure can be formally expressed as: 
\begin{equation}
\mathcal{D}_i^M = g(\mathcal{M}_i; \theta_D) 
\end{equation}
where \(g({\cdot; \theta_D})\) encompasses both the reshaping operations and the degradation network parameterized by \(\theta_D\). The parameter update process for \(\theta_{D}\) is as follows:
\begin{equation} \label{eq:DAM}
\theta^{(k+1)}_{D} = \arg\min_{\theta^{(k)}_{D}} \| \mathcal{L}_i^P - \mathcal{D}_i^M \|_2^2
\end{equation}
where \(\theta^{(k+1)}_{D}\) represents the parameters at the \(k\)-th iteration, and \(\theta^{(k)}_{D}\) denotes the updated parameters after this. 
The optimization minimizes the \(\ell_2\) norm between the \(\mathcal{L}_i^P\) and the network's prediction \(\mathcal{D}_i^M\), effectively guiding the network to restore spectral details lost during degradation.

\begin{figure}[t]
   \centering
   \includegraphics[width=\textwidth]{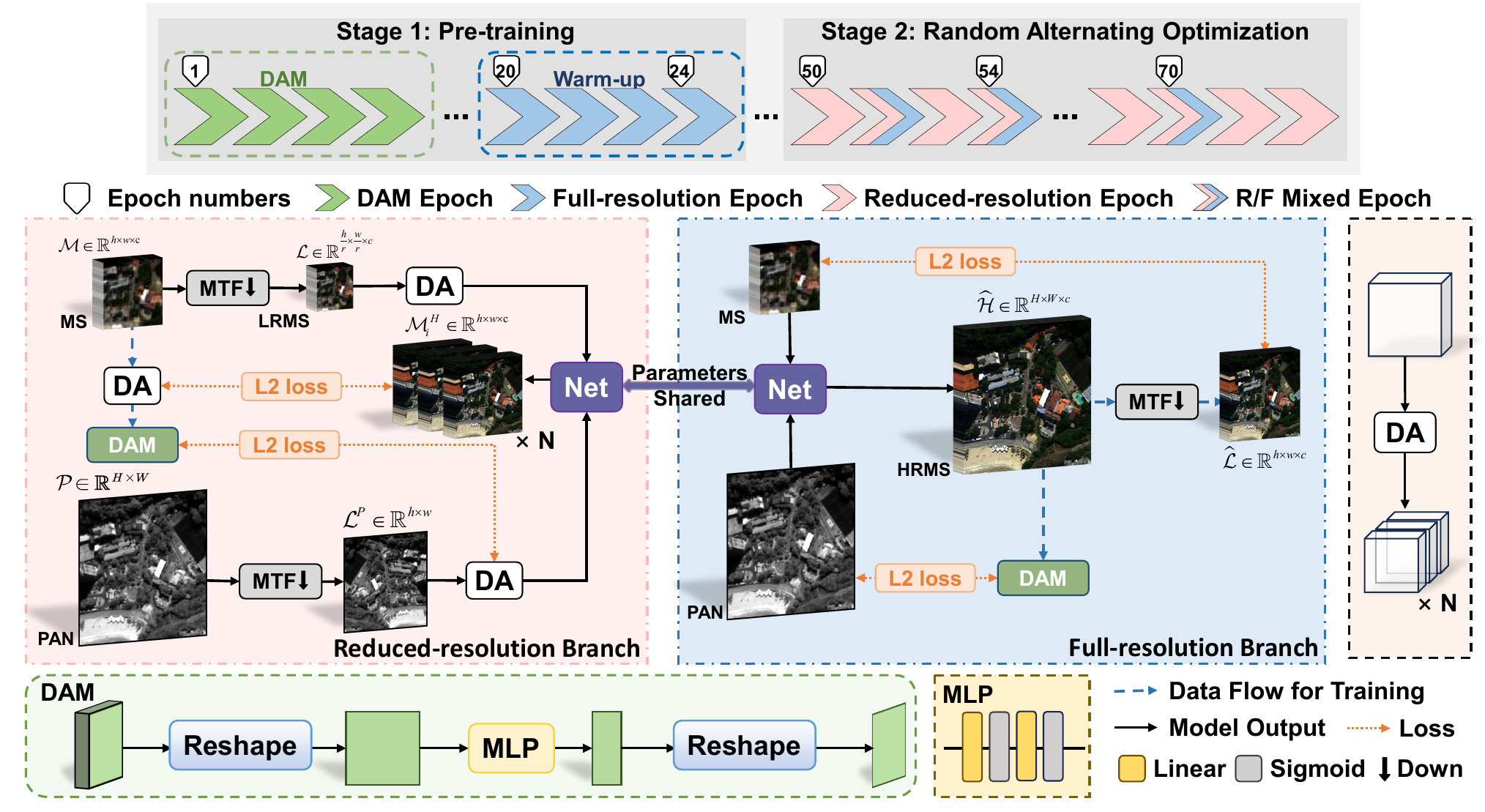}
   \caption{The TRA-PAN training framework. The process, illustrated by the top progress bar, initiates with DAM in the pre-training stage, whose network parameters \(\theta_D\) are subsequently utilized by the RAO stage. Following DAM, a warm-up procedure operates exclusively on the full-resolution branch. Finally, the RAO stage performs random alternating training. The ``R/F Mixed Epoch'' is a key part, where the model concurrently trains on both reduced- and full-resolution branches. This process utilizes data from a single input MS/PAN pair to progressively refine the fusion model.}
   \vspace{-8pt}
   \label{fig:main}
\end{figure}

\paragraph{Warm-up Procedure} The learning process for the fusion network parameters \(\theta_{R}\) begins with this procedure. Specifically, the fusion network is trained exclusively on full-resolution data for \(m\) epochs. Details regarding the full-resolution training are provided in \cref{sec:full}. This procedure aims to reduce the distribution gap between simulation images and real-world data, allowing the model to discover the general region of optimal solutions more effectively. Training initially at full resolution departs from conventional methods, which typically involve pre-training on reduced-resolution data followed by full-resolution fine-tuning. For further details on the impact of varying warm-up epochs and strategies, refer to \cref{exp:warmup_strategy} and \cref{exp:warmup_epochs}.

\subsection{Stage 2: Random Alternating Optimization}
\paragraph{RAO Stage} The Random Alternating Optimization (RAO) stage trains the fusion network using a random alternating strategy. Each epoch invariably involves reduced-resolution training, followed by a probabilistic decision (with probability \(p\)) to incorporate full-resolution training. This injects randomness, acting as a regularizer to mitigate overfitting to a single resolution.
\paragraph{Reduced-resolution} For reduced-resolution training, a small dataset is created from a \(\mathcal{L}^P/\mathcal L\) using the data augmentation from the DAM procedure. The fusion backbone can be any pansharpening architecture. Our experiments adopt FusionNet~\cite{FusionNet} as the base model. The output of this training phase is a fused image \(\mathcal{M}_i^H\). The parameter \(\theta_R\) update process for the fusion network is as follows:
\begin{equation} \label{eq:reduced}
\theta^{(k+1)}_{R} = \arg\min_{\theta^{(k)}_{R}} \| \mathcal{M}_i^H- \mathcal{M}_i \|_2^2
\end{equation}
The parameters \(\theta_{R}\) are iteratively updated by minimizing the \(\ell_2\) norm between \(\mathcal{M}_i^H\) and \(\mathcal{M}_i\). Consequently, the network's performance is enhanced, providing crucial auxiliary support for the overall RAO stage.

\paragraph{Full-resolution}
\label{sec:full}
In full-resolution training, the DAM-trained degradation model is used. Specifically, the predicted HRMS image \(\widehat{\mathcal{H}}\ \in \mathbb{R}^{H \times W \times c}\) is fed into the DAM to generate \(\mathcal{D}^H \in \mathbb{R}^{H \times W}\). 
Spectral loss \(\mathcal{L}_{spectral}\) between \(\mathcal{D}^H\) and \(\mathcal{P}\) recovers high-frequency details, attenuated by the sensor's MTF.
\begin{equation}
\mathcal{L}_{spectral} = \| \mathcal{D}^H - \mathcal{P} \|_2^2
\end{equation}
Additionally, the predicted HRMS image \(\widehat{\mathcal{H}}\) is downsampled to produce \(\widehat{\mathcal L} \in \mathbb{R}^{h \times w \times c}\). To maintain spatial consistency, a spatial loss \(\mathcal{L}_{spatial}\), is introduced. This loss ensures the fusion process preserves essential spatial details and structural integrity. The spatial loss is formulated as:
\begin{equation}
\mathcal{L}_{spatial} = \| \widehat{\mathcal L} - \mathcal{M} \|_2^2
\end{equation}
\begin{table}[t]
\centering
\caption{Comparisons on WV3, QB, and GF2 full-resolution datasets, each with 20 samples. Best: \textbf{bold}, and second-best:\underline{underline}.}
\begin{tabular}{cc@{\hskip 3pt}c@{\hskip 3pt}cc@{\hskip 3pt}c@{\hskip 3pt}cc@{\hskip 3pt}c@{\hskip 3pt}c}
\hline
\multirow{2}{*}{\textbf{Methods}} & \multicolumn{3}{c}{\textbf{ WV3 }} & \multicolumn{3}{c}{\textbf{QB}} & \multicolumn{3}{c}{\textbf{GF2}} \\
\cmidrule(lr){2-4}\cmidrule(lr){5-7} \cmidrule(lr){8-10}
& \textbf{D$_\lambda\downarrow$ }& \textbf{D$_s\downarrow$}& \textbf{HQNR$\uparrow$}& \textbf{D$_\lambda\downarrow$}& \textbf{D$_s\downarrow$}& \textbf{HQNR$\uparrow$}& \textbf{D$_\lambda\downarrow$}& \textbf{D$_s\downarrow$}& \textbf{HQNR$\uparrow$} \\ \hline
TV~\cite{TV}$^{\textcolor{gray}{2013}}$  & 0.024& 0.039& 0.938& 0.055& 0.100& 0.850& 0.055& 0.112& 0.839\\
MTF-GLP-FS ~\cite{MTF-GLP-FS} $^{\textcolor{gray}{2018}}$ & \underline{0.020}& 0.063& 0.918& 0.047& 0.150& 0.810& 0.035& 0.143& 0.828\\
BDSD-PC~\cite{BDSD-PC}$^{\textcolor{gray}{2019}}$  & 0.063& 0.073& 0.869& 0.198& 0.164& 0.672& 0.076& 0.155& 0.781\\
\hline
FusionNet~\cite{FusionNet} $^{\textcolor{gray}{2021}}$ & 0.025& 0.044& 0.932& 0.057& 0.052& 0.894& 0.035& 0.101& 0.867\\
LAGNet~\cite{LAGConv} $^{\textcolor{gray}{2022}}$ & 0.036& 0.041& 0.925& 0.086& 0.068& 0.852& 0.028& 0.079& 0.895\\
LGPNet~\cite{LGPConv}$^{\textcolor{gray}{2023}}$  & 0.022& 0.039& 0.940& 0.074& 0.061& 0.870& 0.030& 0.080& 0.892\\
CANNet~\cite{Content-Adaptive} $^{\textcolor{gray}{2024}}$ & \underline{0.020}& \underline{0.029}& 0.951& \textbf{0.038}& 0.047& \underline{0.917}& \textbf{0.019}& 0.063& 0.919\\
ZS-Pan~\cite{Zero-shot} $^{\textcolor{gray}{2024}}$ &0.026& 0.031& 0.945& 0.053& 0.082& 0.870& 0.054& 0.100& 0.852\\
PanMamba~\cite{Pan-Mamba}$^{\textcolor{gray}{2025}}$  & \underline{0.020}& 0.031& \underline{0.954}& 0.049& \underline{0.044}& 0.910& \underline{0.023}& \underline{0.057}& \underline{0.921}\\
\textbf{Proposed} & \textbf{0.019}& \textbf{0.015}& \textbf{0.966}& \underline{0.045}& \textbf{0.033}& \textbf{0.924}& 0.031& \textbf{0.040}& \textbf{0.930}\\
\hline
\end{tabular}
\vspace{-12pt}
\label{tab:SOTA}
\end{table}
Throughout this process, we continue to optimize the parameter \(\theta_{R}\) used in the reduced-resolution phase. The whole optimization process in the RAO stage is formalized as:
\begin{equation} \label{eq:full}
\theta^{(k+1)}_{R} = \arg\min_{\theta^{(k)}_{R}}(\lambda_1 \mathcal{L}_{spectral}(\widehat{\mathcal{D}^P}, \mathcal{P}) + \lambda_2 \mathcal{L}_{spatial}(\widehat{\mathcal L}, \mathcal{M}))
\end{equation}
where \(\theta_{R}\) represents the parameters learned during the RAO stage, and \(\lambda_1\) and \(\lambda_2\) are hyperparameters used to balance the importance of the spectral and spatial losses.
\section{Experiments\label{checklist:Experimental}}
\label{sec:experiments}
Our experiments utilize three satellite imagery datasets—WorldView-3 (WV3), QuickBird (QB), and GaoFen-2 (GF2)—sourced from the PanCollection repository~\cite{dengMachineLearningPansharpening2022}\footnote{\url{https://github.com/liangjiandeng/PanCollection}}. Wald's protocol~\cite{waldFusionSatelliteImages1997, dengDetailInjectionBasedDeep2021} is applied for MTF filtering. Due to our method's instance-specific nature, cross-dataset generalization experiments on WorldView-2 (WV2) are omitted. Performance in real-world scenarios is assessed using established full-resolution metrics: \(D_s\), \(D_\lambda\), and HQNR~\cite{arienzo2022full}. HQNR, a comprehensive metric derived from \(D_s\) and \(D_\lambda\), is prioritized for its widespread adoption in assessing fusion quality. We retain HQNR as the primary fusion quality measure for consistency and robust comparisons with prior studies. Although newer metrics with potentially better Human Visual System (HVS) correlation exist~\cite{scarpa2022full, agudelo2019perceptual, stkepien2022no}, they are not yet broadly adopted by the research community. All model loss functions are formulated using \(\ell_2\) norm, and the Adam optimizer~\cite{Adam} is employed throughout all stages. For the RAO stage, training is conducted with a batch size of 8 over 250 epochs. The DAM procedure is trained using a batch size of 1 and for 250 epochs. Experiments are primarily performed on an NVIDIA GeForce RTX 4070 Ti SUPER GPU. \textit{Due to the space limitation, additional experiments including the impact of various backbones, analysis of sampling ratio \(p\), and further training details are provided in the supplementary material.}

\begin{figure}[t]
   \centering
   \includegraphics[width=\textwidth]{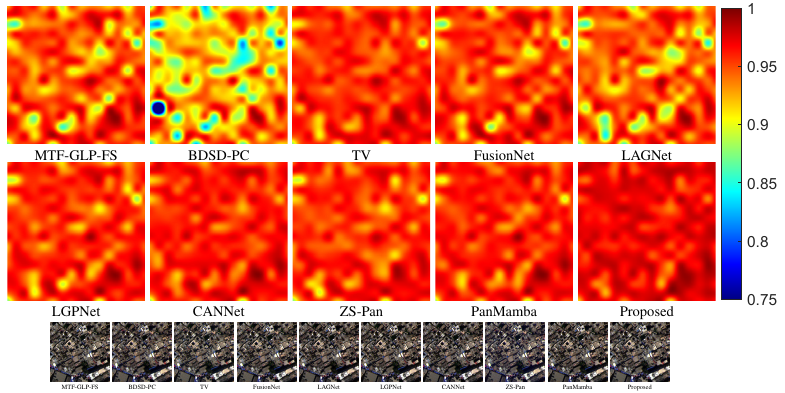}
   \caption{The HQNR maps (Top) and visual results (bottom) of all compared approaches on the WV3 full-resolution dataset.}
   \vspace{-15pt}
\end{figure}

\subsection{Comparison with SOTA methods}
We benchmark our method against nine competitors: three classical (MTF-GLP-FS~\cite{MTF-GLP-FS}, BDSD-PC~\cite{BDSD-PC}, TV~\cite{TV}) and six deep learning (DL)-based (FusionNet~\cite{FusionNet}, LAGNet~\cite{LAGConv}, LGPNet~\cite{LGPConv}, CANNet~\cite{Content-Adaptive}, ZS-Pan~\cite{Zero-shot}, PanMamba~\cite{Pan-Mamba}). Because of representativeness, we selected these methods from reduced-resolution training and hybrid training approaches. This comparison showcases our method's generalization, aiming for superior performance with less training and fewer parameters, and its advantages over phase-based methods like ZS-Pan.
FusionNet is our fusion network that reduces training time and mitigates overfitting. Since our method is designed for real-world scenarios, the comparative analysis primarily focuses on full-resolution results. \textit{We also provide reduced-resolution metrics and visuals in the supplementary material.} \cref{tab:SOTA} provides a comprehensive comparison with SOTA techniques across three datasets. Our approach demonstrates strong performance across all metrics, particularly on the WV3 dataset. Specifically, it achieves HQNR improvements of 0.012 (WV3), 0.007 (QB), and 0.009 (GF2) over the second-best methods. These results underscore our framework's distinct advantages and competitiveness.

\vspace{-4pt}
\subsection{Ablation Study and Analysis\label{checklist:4.2}}
\setlength{\columnsep}{\OriginalColumnsep}
\vspace{-4pt}
\paragraph{Warm-up Strategy}
\label{exp:warmup_strategy}
\begin{wraptable}{r}{0.5\textwidth}
\vspace{-13pt}
\centering
\caption{Experiment on warm-up strategies at different resolutions.}
\setlength{\tabcolsep}{4pt} 
\begin{tabular}{cccc}
\hline
\multirow{2}{*}{\textbf{Warmup Strategy}} & \multicolumn{3}{c}{\textbf{WV3}}\\
\cmidrule(lr){2-4}
& \textbf{D$_\lambda \downarrow$} & \textbf{D$s$$\downarrow$} & \textbf{HQNR$\uparrow$} \\
\hline
w Full & 0.0193 & 0.0154 & 0.9657 \\
w Reduced & 0.0286 & 0.0252 & 0.9469 \\
w/o Warmup   & 0.0266 & 0.0263 & 0.9478 \\
\hline
\end{tabular}
\label{table:warmup_strategy}
\vspace{-8pt}
\end{wraptable}
To demonstrate the effectiveness of the proposed warm-up strategy, an ablation study is conducted on the WV3 dataset, comparing three approaches: 20 epochs of full-resolution warm-up, 20 epochs of reduced-resolution warm-up, and no warm-up. As indicated in \cref{table:warmup_strategy}, full-resolution warm-up (first row) significantly aids the network in finding optimal solutions. Conversely, reduced-resolution warm-up (second row) leads to a performance decrease compared to no warm-up (third row), confirming its supplementary role. Our experiments reveal that full-resolution warm-up yields superior results, as direct training on reduced-resolution data often traps the network in suboptimal solutions, especially with real-world data.

\vspace{-8pt}
\paragraph{Training Order}
\begin{wraptable}{r}{0.5\textwidth}
\vspace{-13pt}
\centering
\caption{Training order strategies for different resolutions.}
\setlength{\tabcolsep}{4pt} 
\begin{tabular}{cccc}
\hline
\multirow{2}{*}{\textbf{Training Order}} & \multicolumn{3}{c}{\textbf{WV3}} \\
\cmidrule(lr){2-4}
& \textbf{D$_\lambda \downarrow$} & \textbf{D$s$$\downarrow$} & \textbf{HQNR$\uparrow$} \\
\hline
Alternating  & 0.0193 & 0.0154 & 0.9657 \\
Full/Reduced  & 0.0316 & 0.0323 & 0.9372 \\
Reduced/Full & 0.0234 & 0.0393 & 0.9384 \\
\hline
\end{tabular}
\label{table:training_order}
\vspace{-8pt}
\end{wraptable}
To validate the effectiveness of our random alternating training, we evaluate three distinct training order strategies, with results presented in \cref{table:training_order}. The first strategy (first row) applies our proposed random alternation, where full-resolution training is chosen with probability \(p=0.8\) and reduced-resolution training with probability \(0.2\) in each epoch. The second strategy (second row) employs the same number of full- and reduced-resolution epochs but in a fixed sequence: full-resolution training followed by reduced-resolution training (denoted ``Full/Reduced''). The third strategy (third row) reverses this fixed sequence, performing reduced-resolution training before full-resolution training. As shown in \cref{table:training_order}, the model's performance significantly deteriorates under the fixed-sequence configurations, thereby underscoring the efficacy of the proposed random alternating training strategy.

\paragraph{Random Sampling Strategy}
\begin{wraptable}{r}{0.65\textwidth}
\centering
\caption{Comparison of different random sampling strategies for training.}
\setlength{\tabcolsep}{4pt} 
\begin{tabular}{cccc}
\hline
\multirow{2}{*}{\textbf{Random Sampling Strategy}} & \multicolumn{3}{c}{\textbf{WV3}} \\
\cmidrule(lr){2-4}
& \textbf{D$_\lambda \downarrow$} & \textbf{D$s$$\downarrow$} & \textbf{HQNR$\uparrow$} \\
\hline
Always Reduced  & 0.0193 & 0.0154 & 0.9657 \\
Always Full  & 0.0248 & 0.0326 & 0.9435 \\
Random  & 0.0256 & 0.0261 & 0.9491 \\
\hline
\end{tabular}
\label{table:random_sampling_strategy}
\vspace{-8pt}
\end{wraptable}
We investigate the impact of different random sampling strategies on performance by testing three configurations. The first configuration ensures reduced-resolution training each epoch, with full-resolution training selected with probability \(p\) (details in \cref{alg:framework}, \cref{alg:line7}-\cref{alg:line12}). The second configuration ensures full-resolution training each epoch, with reduced-resolution training selected with probability \(p\), which involves swapping the procedures detailed in \cref{alg:line7}-\cref{alg:line12} with those in \cref{alg:line13}-\cref{alg:line18}. The third configuration randomly selects either reduced- or full-resolution training for each epoch, achievable by adding a sub-condition ``branch is REDUCED'' after \cref{alg:line18}. Results \cref{table:random_sampling_strategy} indicate that continuous full-resolution training can lead to optimization drift due to weak supervision. The first strategy, adopted in our SOTA experiments, outperformed the others, highlighting the detrimental effect of excessive continuous full-resolution training.

\vspace{-6pt}
\subsection{Discussion\label{checklist:4.3}}

\begin{figure}[t]
    \centering
    \includegraphics[width=\textwidth]{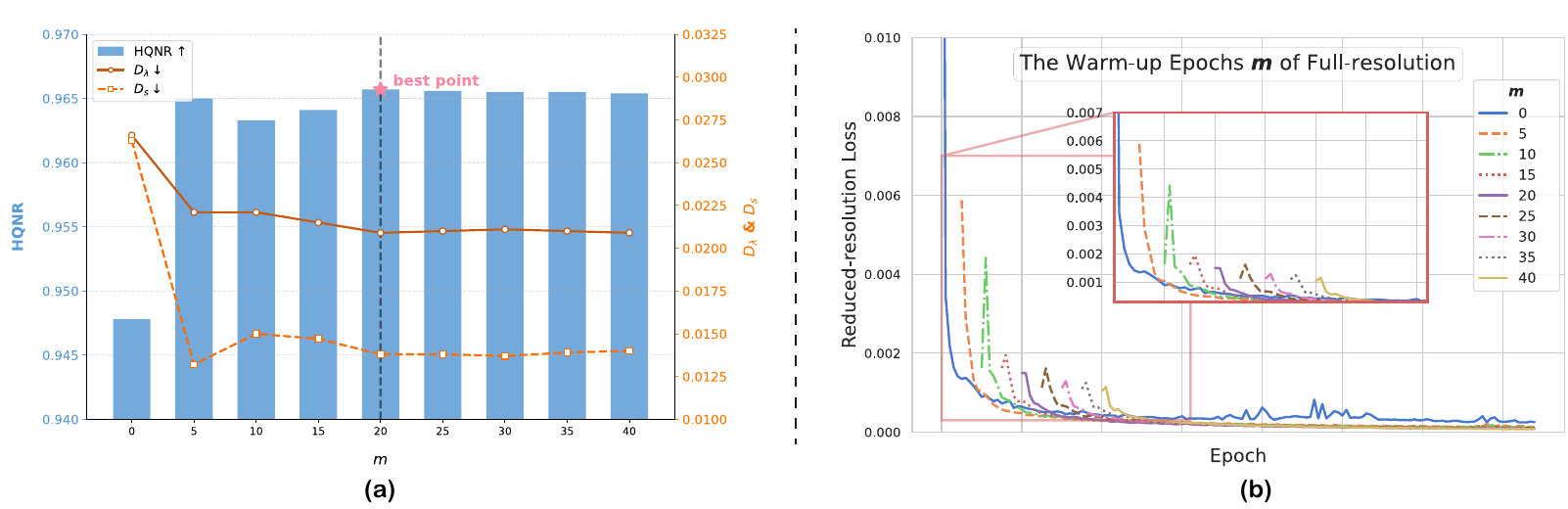}
    \caption{Impact of warm-up epochs \(m\) on model performance and training dynamics. The left panel illustrates the effect of varying warm-up epochs on the HQNR, alongside \(D_\lambda\), \(D_s\) metrics. The right panel depicts the relationship between the number of \(m\) and the corresponding loss of the fusion network during the subsequent reduced-resolution training in the RAO stage.}
    \vspace{-14pt}
    \label{fig:warmup}
\end{figure}

\vspace{-3pt}
\paragraph{Warm-up Epochs}
\label{exp:warmup_epochs}
This section investigates the influence of the warm-up epoch count \(m\), a critical hyperparameter. Experiments on the WV3 dataset, holding other hyperparameters constant, show that increasing \(m\) from 0 to 20 improves the model's ability to find optimal solutions during full-resolution training (\cref{fig:warmup} (a)). Notably, \cref{fig:warmup} (a) also illustrates that the fusion quality is not sensitive to \(m\) within a reasonable range, demonstrating the robustness of our method. This reduces initial loss in subsequent reduced-resolution training, accelerates convergence, and increases HQNR. At \(m=20\), the model achieves an optimal initial state for the RAO stage. However, larger \(m\) values risk trapping the model in local optima during full-resolution training, diminishing the corrective effect of reduced-resolution training, and slightly reducing HQNR.

The line graph in \cref{fig:warmup} (b) illustrates the relationship between warm-up epochs and network loss during subsequent reduced-resolution training. This highlights the criticality of \(m\): too few epochs prevent reaching the optimal solution region, while too many hinder adaptation to reduced-resolution data, leading to suboptimal performance. Thus, selecting an appropriate \(m\) is crucial for a robust RAO stage starting point.
\vspace{-4pt}
\section{Conclusion and Limitation}
\vspace{-4pt}
\label{sec:conclusion}
This paper introduces TRA-PAN, a two-stage one-shot pansharpening framework using random alternating training alongside a warm-up procedure, enabling instance-specific optimization for any given MS/PAN image pair. Alternating training optimizes the fusion model, while warm-up accelerates convergence. Experiments show that TRA-PAN outperforms SOTA methods regarding metrics and visual quality, proving its practical effectiveness. \label{checklist:limit}A limitation is slightly increased training time due to random alternation, though it remains acceptable.

\begin{ack}
\end{ack}

{\small
\bibliographystyle{ieeenat_fullname}
\bibliography{11_references}
}




\newpage
\section*{NeurIPS Paper Checklist}
\begin{enumerate}

\item {\bf Claims}
    \item[] Question: Do the main claims made in the abstract and introduction accurately reflect the paper's contributions and scope?
    \item[] Answer: \answerYes{} 
    \item[] Justification: The abstract and introduction clearly outline the main contributions and scope, including the TRA-PAN framework, its two-stage process (Pre-training, RAO), and its instance-specific optimization for pansharpening. The described methodology and experimental setup support the claims of robust, high-quality fusion and superior performance.

\item {\bf Limitations}
    \item[] Question: Does the paper discuss the limitations of the work performed by the authors?
    \item[] Answer: \answerYes{} 
    \item[] Justification: The paper discusses its limitations in \cref{checklist:limit}.

\item {\bf Theory assumptions and proofs}
    \item[] Question: For each theoretical result, does the paper provide the full set of assumptions and a complete (and correct) proof?
    \item[] Answer: \answerNA{} 
    \item[] Justification: This paper does not contain rigorous theoretical assumptions and proofs. However, an approximate analysis of the Degradation-Aware Modeling (DAM) capability to capture spectral degradation is provided in the supplementary material.

    \item {\bf Experimental result reproducibility}
    \item[] Question: Does the paper fully disclose all the information needed to reproduce the main experimental results of the paper to the extent that it affects the main claims and/or conclusions of the paper (regardless of whether the code and data are provided or not)?
    \item[] Answer: \answerYes{} 
    \item[] Justification: The paper provides significant details regarding the experimental setup in \cref{checklist:Experimental}, including datasets used (WV3, QB, GF2 from PanCollection repository), MTF filtering protocol, evaluation metrics, optimizer details (Adam), batch sizes, epoch numbers, and GPU used. Further information like network backbone specifics, MLP architecture for DAM, data augmentation procedure, MTF filter specifics, and analysis of sampling ratio \(p\) are stated in the \textit{supplementary material}. The complete code will be published on GitHub upon acceptance for further research and discussion.

\item {\bf Open access to data and code}
    \item[] Question: Does the paper provide open access to the data and code, with sufficient instructions to faithfully reproduce the main experimental results, as described in supplemental material?
    \item[] Answer: \answerYes{} 
    \item[] Justification: The datasets are publicly accessible, and the code will be released upon acceptance.

\item {\bf Experimental setting/details}
    \item[] Question: Does the paper specify all the training and test details (e.g., data splits, hyperparameters, how they were chosen, type of optimizer, etc.) necessary to understand the results?
    \item[] Answer: \answerYes{} 
    \item[] Justification: \cref{checklist:Experimental} details datasets, evaluation metrics, optimizer, batch sizes, and epoch numbers. \cref{checklist:4.2} and \cref{checklist:4.3} discuss hyperparameter choices like warm-up epochs. The supplementary material contains further details on backbones, sampling ratio \(p\), and training.

\item {\bf Experiment statistical significance}
    \item[] Question: Does the paper report error bars suitably and correctly defined or other appropriate information about the statistical significance of the experiments?
    \item[] Answer: \answerNo{} 
    \item[] Justification: The quantitative results presented in \cref{tab:SOTA} report single numerical values for the metrics and do not include error bars, standard deviations, or other measures of statistical significance for the experimental results.

\item {\bf Experiments compute resources}
    \item[] Question: For each experiment, does the paper provide sufficient information on the computer resources (type of compute workers, memory, time of execution) needed to reproduce the experiments?
    \item[] Answer: \answerYes{} 
    \item[] Justification: \cref{checklist:Experimental} states that experiments were primarily performed on an NVIDIA GeForce RTX 4070 Ti SUPER GPU. Supplementary material further specifies.
    
\item {\bf Code of ethics}
    \item[] Question: Does the research conducted in the paper conform, in every respect, with the NeurIPS Code of Ethics \url{https://neurips.cc/public/EthicsGuidelines}?
    \item[] Answer: \answerYes{}
    \item[] Justification: The research adheres to the ethical guidelines set forth by NeurIPS, ensuring responsible conduct in all aspects of the study.

\item {\bf Broader impacts}
    \item[] Question: Does the paper discuss both potential positive societal impacts and negative societal impacts of the work performed?
    \item[] Answer: \answerYes{}
    \item[] Justification: \cref{checklist:impact} details positive impacts by improving high-resolution multispectral imagery for crucial remote sensing applications, such as environmental monitoring, urban planning, and precision agriculture, enhancing practical utility and robustness. 
    
\item {\bf Safeguards}
    \item[] Question: Does the paper describe safeguards that have been put in place for responsible release of data or models that have a high risk for misuse (e.g., pretrained language models, image generators, or scraped datasets)?
    \item[] Answer: \answerNA{}
    \item[] Justification: The paper does not describe specific release safeguards as its pansharpening framework utilizes existing public datasets and is not considered to release new, high-risk assets.

\item {\bf Licenses for existing assets}
    \item[] Question: Are the creators or original owners of assets (e.g., code, data, models), used in the paper, properly credited and are the license and terms of use explicitly mentioned and properly respected?
    \item[] Answer: \answerYes{} 
    \item[] Justification: The paper credits the PanCollection repository~\cite{dengMachineLearningPansharpening2022}, accessible via \url{https://github.com/liangjiandeng/PanCollection}, for the datasets utilized. This repository is licensed under the GNU General Public License v3.0 (GPL-3.0).

\item {\bf New assets}
    \item[] Question: Are new assets introduced in the paper well documented and is the documentation provided alongside the assets?
    \item[] Answer: \answerNA{}
    \item[] Justification: The paper does not release new assets.

\item {\bf Crowdsourcing and research with human subjects}
    \item[] Question: For crowdsourcing experiments and research with human subjects, does the paper include the full text of instructions given to participants and screenshots, if applicable, as well as details about compensation (if any)? 
    \item[] Answer: \answerNA{}
    \item[] Justification: The paper does not involve crowdsourcing or research with human subjects.

\item {\bf Institutional review board (IRB) approvals or equivalent for research with human subjects}
    \item[] Question: Does the paper describe potential risks incurred by study participants, whether such risks were disclosed to the subjects, and whether Institutional Review Board (IRB) approvals (or an equivalent approval/review based on the requirements of your country or institution) were obtained?
    \item[] Answer: \answerNA{} 
    \item[] Justification: The paper does not involve crowdsourcing or research with human subjects.

\item {\bf Declaration of LLM usage}
    \item[] Question: Does the paper describe the usage of LLMs if it is an important, original, or non-standard component of the core methods in this research? Note that if the LLM is used only for writing, editing, or formatting purposes and does not impact the core methodology, scientific rigorousness, or originality of the research, declaration is not required.
    \item[] Answer: \answerNA{} 
    \item[] Justification: The paper has nothing to do with using Large Language Model (LLM) as a central component of its research methodology.

\end{enumerate}
\end{document}